\documentclass[11pt,a4paper]{article}
\usepackage[hyperref]{emnlp-ijcnlp-2019}
\usepackage{times}
\usepackage{latexsym}

\usepackage{url}
\usepackage{hyperref}
\usepackage{booktabs}
\usepackage{multirow}
\usepackage{graphicx}
\usepackage{amsmath}
\usepackage{mathtools}
\usepackage{amssymb}
\usepackage{amsthm}
\usepackage{amsfonts}
\usepackage{csquotes}
\usepackage{breqn}
\usepackage{subfig}
\usepackage{siunitx}

\usepackage{enumitem}
\usepackage{paralist}
\usepackage{soul}
\usepackage{footmisc}
\usepackage{boldline}
\usepackage{makecell}
\usepackage[normalem]{ulem}
\useunder{\uline}{\ul}{}

\aclfinalcopy 

\title{Generating a Common Question from Multiple Documents using Multi-source Encoder-Decoder Models}

\author{Woon Sang Cho$^{\Diamond}$\quad Yizhe Zhang$^{\dagger}$\quad Sudha Rao$^{\dagger}$\quad Chris Brockett$^{\dagger}$\quad Sungjin Lee$^{\S}$\thanks{\hspace{0.1in}Work was done when affiliated with Microsoft Research AI}\\
  $^{\Diamond}$Princeton University \\
  $^{\dagger}$Microsoft Research AI \\
  $^{\S}$Amazon Alexa AI \\
  {\tt $^\Diamond$\{woonsang\}@princeton.edu}, {\tt $^\S$\{sungjinl\}@amazon.com}\\
  {\tt $^\dagger$\{yizzhang,sudhra,chrisbkt\}@microsoft.com}
}

\date{}

\begin{document}
\maketitle
\begin{abstract}
Ambiguous user queries in search engines result in the retrieval of documents that often span multiple topics. One potential solution is for the search engine to generate multiple refined queries, each of which relates to a subset of the documents spanning the same topic. A preliminary step towards this goal is to generate a question that captures common concepts of multiple documents. We propose a new task of generating common question from multiple documents and present simple variant of an existing multi-source encoder-decoder framework, called the Multi-Source Question Generator (MSQG). We first train an RNN-based single encoder-decoder generator from (single document, question) pairs. At test time, given multiple documents, the \textit{Distribute} step of our MSQG model predicts target word distributions for each document using the trained model. The \textit{Aggregate} step aggregates these distributions to generate a common question. This simple yet effective strategy significantly outperforms several existing baseline models applied to the new task when evaluated using automated metrics and human judgments on the MS-MARCO-QA dataset.

\end{abstract}

\section{Introduction}\label{intro}
             
Search engines return a list of results in response to a user query.
In the case of ambiguous queries, retrieved results often span multiple topics and might benefit from further clarification from the user. 
One approach to disambiguate such queries is to first partition the retrieved results by topic and then ask the user to choose from queries refined for each partition. 

For example, a query \textit{`how good is apple?'} could retrieve documents some of which relate to apple the fruit, and some of which relate to Apple the company. In such a scenario, if the search engine generates two refinement queries \textit{`how good is apple the fruit?'} and \textit{`how good is the company apple?'}, the user could then choose one of it as a way to clarify her initial query. 




In this work, we take a step towards this aim by proposing a model that generates a common question that is relevant to a set of documents.
At training time, we train a standard sequence-to-sequence model \cite{seq2seq} with a large number of (single document, question) pairs to generate a relevant question given a single document.
At test time, given multiple ($N$) input documents, we use our model, called the Multi-Source Question Generator (MSQG), to allow document-specific decoders to collaboratively generate a common question. 
We first encode the $N$ input documents separately using the trained encoder. Then, we perform an iterative procedure to i) (Distribute step) compute predictive word distributions from each document-specific decoder based on previous context and generation ii) (Aggregate step) aggregate predictive word distributions by voting and generate a single shared word for all decoders. These two steps are repeated until an end-of-sentence token is generated. We train and test our model on the MS-MARCO-QA dataset and evaluate it by assessing whether the original passages can be retrieved from the generated question, as well as human judgments for fluency, relevancy, and answerability. Our model significantly outperforms multiple baselines. Our main contributions are: 
\begin{compactitem}
    \item[i)] a new task of generating a common question from multiple documents, where a common question target is does \textit{not} exist, unlike multi-lingual sources to common language translation tasks.
	\item[ii)] an extensive evaluation of an existing multi-source encoder-decoder models including our simple variant model for generating a common question.
	\item[iii)]  an empirical evaluation framework based on automated metrics and human judgments on \textit{answerability}, \textit{relevancy}, and \textit{fluency} to extensively evaluate our proposed MSQG model against the baselines.
\end{compactitem}


\section{Related Work}\label{relatedwork}

The use of neural networks to generate natural language questions has mostly focused on question answering \cite{labutov-etal-2015-deep,serban-etal-2016-generating,a92dd6bf37bd48f99a691c3ea8343823,song2017unified,duan-etal-2017-question,du-etal-2017-learning,buck2017ask,song-etal-2018-leveraging,harrison2018neural,sun-etal-2018-answer}. A number of works process multiple passages by concatenating, adding, or attention-weight-summing among passage features into a single feature, and use it for downstream tasks \citep{zoph-knight-2016-multi,garmash-monz-2016-ensemble,libovicky-helcl-2017-attention,wang-etal-2018-multi-passage,DBLP:journals/corr/abs-1811-11374,lebanoff-etal-2018-adapting,celikyilmaz-etal-2018-deep,nishimura-etal-2018-multi,libovicky-etal-2018-input,DBLP:journals/corr/abs-1811-04897,DBLP:journals/corr/abs-1901-02262}. Our processing mechanisms are similar to \citeauthor{garmash-monz-2016-ensemble} \shortcite{garmash-monz-2016-ensemble}, \citeauthor{DBLP:journals/corr/FiratSAYC16} \shortcite{DBLP:journals/corr/FiratSAYC16}, and \citeauthor{dong-smith-2018-multi} \shortcite{dong-smith-2018-multi}. The information retrieval literature is primarily concerned with reformulating queries, by either selecting expansion terms from candidates as in pseudo-relevance feedback \citep{Salton:1971:SRS:1102022,Zhai:2001:MFL:502585.502654,Xu:1996:QEU:243199.243202,Metzler:2007:LCE:1277741.1277796,Cao:2008:SGE:1390334.1390377,bernhard-2010-query, nogueira-cho-2017-task,li-etal-2018-nprf}. Our task differs because there is no \textit{supervision} unlike multi-lingual translation tasks where a single target translation is available given sources from multiple languages.


\section{Method}\label{model}


\subsection{Multi-Source Question Generator}
Our Multi-Source Question Generator (MSQG) model introduces a mechanism to generate a common question given multiple documents. At \textit{training} time, it employs a standard sequence-to-sequence (S2S) model using a large number of (single document, question) pairs.
At \textit{test} time, it generates a common question given multiple documents, similar to \citeauthor{garmash-monz-2016-ensemble} \shortcite{garmash-monz-2016-ensemble} and \citeauthor{DBLP:journals/corr/FiratSAYC16} \shortcite{DBLP:journals/corr/FiratSAYC16}. Specifically, our MSQG model iterates over two interleaved steps, until an end-of-sentence (EOS) token is generated:
\paragraph{Distribute Step}
During the Distribute step, we take an instance of the trained S2S model, and perform inference with $N$ different input documents.
Each document is then encoded using one copy of the model to generate a unique target vocabulary distribution $\mathcal{P}^{\text{dec}}_{i,t}$ (for document $i$, at time $t$) for the next word. Note that source information comes from not only encoded latent representation from a source document, but also the cross-attention between source and generation. 

\paragraph{Aggregate Step}
During the Aggregate step, we aggregate the $N$ different target distributions into one distribution by averaging them as below: 
$$\mathcal{\tilde{P}}^{\text{dec}}_t = \frac{1}{N}\big(\beta_1\mathcal{P}^{\text{dec}}_{1,t} + \beta_2\mathcal{P}^{\text{dec}}_{2,t} + \cdots + \beta_{N}\mathcal{P}^{\text{dec}}_{N,t}  \big)$$ where $\mathcal{\tilde{P}}^{\text{dec}}_t$ is the final decoding distribution at time $t$, and $\Sigma_i^N\beta_i=N$. 
In our experiments, we weight all the decoding distributions equally ($\beta_i=1$) to smooth out features that are distinct in each document $i$, where $i \in \{1,\dots, N\}$.

Note that the \textit{average} Aggregate can be perceived as a \textit{majority voting} scheme, in that each document-specific decoder will vote over the vocabulary and the final decision is made in a collaborative manner. 
We also experimented with different Aggregate functions: $(i)$ MSQG$^{mult}$ 
multiplies the distributions, which is analogous to a \textit{unanimous voting} scheme. However, it led to sub-optimal results since one unfavorable distribution can discourage decoding of certain common words. 
$(ii)$ MSQG$^{max}$ takes the maximum probability of each word across $N$ distributions and normalizes them into a single distribution, but it could not generate sensible questions so we excluded from our pool of baselines. 

\begin{figure}[t!]
\centering
\includegraphics[width=1.0\linewidth]{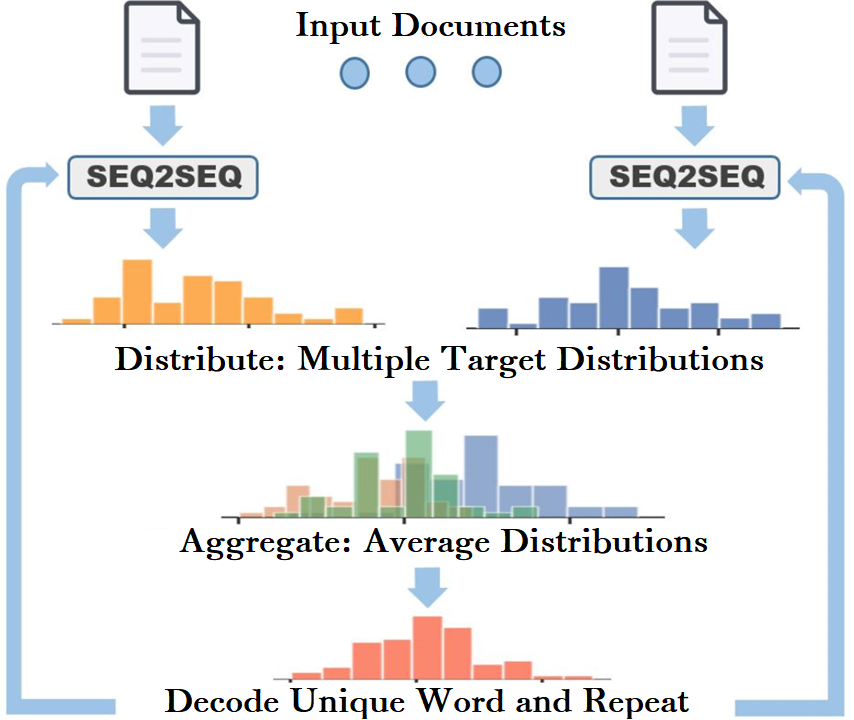} 
\caption{Multi-Source Question Generator (MSQG) model at test time. The simple architecture significantly outperforms the baselines for generating common questions, based on a number of metrics.}
\label{MSQG}
\end{figure}

\subsection{Model Variants}
\paragraph{Avoiding repetitive generation} We observed that naively averaging the target distributions at every decoding time continually emphasized the common topic, thereby decoding repetitive topic words. 
To increase the diversity of generated tokens, we mask those tokens that have already been decoded in subsequent decoding steps. 
This strategy is reasonable for our task since questions generally tend to be short and rarely have repeated words. This mechanism can be viewed as a hard counterpart of the coverage models developed in \citeauthor{tu-etal-2016-modeling} \shortcite{tu-etal-2016-modeling} and  \citeauthor{see-etal-2017-get} \shortcite{see-etal-2017-get}. We denote this feature by $rmrep$ in subscript.

\paragraph{Shared encoder feature} To initialize multiple decoders with the common meaning of the documents in a partition, we broadcast the \textit{mean} of encoded latent representation
to each decoder and denote this variant by the subscript $sharedh$. 
Note that the source document can affect the generated target vocabulary distribution $\mathcal{P}^{\text{dec}}_{i,t}$ at Distribute step through source-generation cross-attention.

\section{Results}\label{experiments}

\subsection{Experimental setup} Our training method uses the standard LSTM-based \citep{Hochreiter:1997:LSM:1246443.1246450} S2S with bi-linear attention \cite{luong-etal-2015-effective}. 
An input to our encoder is a concatenation of 100-dim GloVe \citep{pennington2014glove} vector, 100-dim predicate location vector, and 1024-dim ELMo \citep{Peters:2018} vector. 
Targets are embedded into 100-dim vectors.
The S2S is bi-directional with a 256-dim bi-linear attention in each direction with ReLU \citep{Nair:2010:RLU:3104322.3104425}. 
Our encoder has two layers and we use an Adam \citep{Kingma2014AdamAM} with a learning rate of $2\mathrm{e}{-5}$. 


\subsection{Baselines}
\paragraph{S2S} We compare our model with a standard S2S baseline where we \textit{concatenate} the $N$ documents into a single document to generate a question. We provide detailed discussions about the effect of document order in supplementary material (SM).
 Two variants are considered (S2S and S2S$_{rmrep}$). Beam size is set to 5.

\paragraph{MESD} 

    We also compare our model with the multi-encoder single-decoder (MESD) baseline where documents are encoded individually into $\{v_i\}_{i=1}^N$. The single decoder's initial hidden state is initialized by the $\text{mean}$ of $\{v_i\}_{i=1}^N$, following \citep{dong-smith-2018-multi}. 

\begin{table*}[]
\centering
\resizebox{\textwidth}{!}{%
\begin{tabular}{ccccccccccccccccc}
\hline
\multicolumn{5}{c}{Fluency} &  & \multicolumn{5}{c}{Relevancy} &  & \multicolumn{5}{c}{Answerability} \\ \cline{1-5} \cline{7-11} \cline{13-17} 
 \multicolumn{2}{c|}{} & MSQG & S2S & Human & & \multicolumn{2}{c|}{} & MSQG & S2S & Human & & \multicolumn{2}{c|}{} & MSQG & S2S & Human \\ \cline{1-5} \cline{7-11} \cline{13-17} 
 
\multicolumn{2}{c|}{Completely grammatical} & \multicolumn{1}{c|}{81.37\%} & \multicolumn{1}{c|}{71.28\%} & \textbf{82.48}\% &  & \multicolumn{2}{c|}{Completely relevant} & \multicolumn{1}{c|}{84.00\%} & \multicolumn{1}{c|}{71.00\%} & \textbf{86.40}\% &  & \multicolumn{2}{c|}{Completely answered} & \multicolumn{1}{c|}{71.89\%} & \multicolumn{1}{c|}{48.45\%} & \textbf{72.79}\% \\

\multicolumn{2}{c|}{Comprehensible} & \multicolumn{1}{c|}{16.58\%} & \multicolumn{1}{c|}{21.81\%} & 16.35\% &  & \multicolumn{2}{c|}{Somewhat relevant} & \multicolumn{1}{c|}{7.06\%} & \multicolumn{1}{c|}{7.56\%} & 6.69\% &  & \multicolumn{2}{c|}{Somewhat answered} & \multicolumn{1}{c|}{7.28\%} & \multicolumn{1}{c|}{5.83\%} & 6.83\% \\

\multicolumn{2}{c|}{Not comprehensible} & \multicolumn{1}{c|}{2.05\%} & \multicolumn{1}{c|}{6.91\%} & 1.17\% &  & \multicolumn{2}{c|}{Not relevant} & \multicolumn{1}{c|}{8.94\%} & \multicolumn{1}{c|}{21.44\%} & 6.91\% &  & \multicolumn{2}{c|}{Not answered} & \multicolumn{1}{c|}{20.83\%} & \multicolumn{1}{c|}{45.72\%} & 20.38\% \\ \cline{1-5} \cline{7-11} \cline{13-17} 
 &  &  &  &  &  &  &  &  &  &  &  &  &  &  &  &  \\
\multicolumn{5}{c}{\textit{Human judges preferred:}} &  & \multicolumn{5}{c}{\textit{Human judges preferred:}} &  & \multicolumn{5}{c}{\textit{Human judges preferred:}} \\ \cline{1-5} \cline{7-11} \cline{13-17} 
\multicolumn{2}{c|}{Our Method} & \multicolumn{1}{c|}{Neutral} & \multicolumn{2}{c}{Comparison} &  & \multicolumn{2}{c|}{Our Method} & \multicolumn{1}{c|}{Neutral} & \multicolumn{2}{c}{Comparison} &  & \multicolumn{2}{c|}{Our Method} & \multicolumn{1}{c|}{Neutral} & \multicolumn{2}{c}{Comparison} \\ \cline{1-5} \cline{7-11} \cline{13-17} 
MSQG & \multicolumn{1}{c|}{\bf{75.77\%}} & \multicolumn{1}{c|}{9.74\%} & 14.49\% & S2S &  & MSQG & \multicolumn{1}{c|}{\bf{79.22\%}} & \multicolumn{1}{c|}{8.06\%} & 12.72\% & S2S &  & MSQG & \multicolumn{1}{c|}{\bf{78.50\%}} & \multicolumn{1}{c|}{9.09\%} & 12.40\% & S2S \\
MSQG & \multicolumn{1}{c|}{42.11\%} & \multicolumn{1}{c|}{10.66\%} & \bf{47.24\%} & Human &  & MSQG & \multicolumn{1}{c|}{40.81\%} & \multicolumn{1}{c|}{9.67\%} & \bf{49.52\%} & Human &  & MSQG & \multicolumn{1}{c|}{40.66\%} & \multicolumn{1}{c|}{10.26\%} & \bf{49.08\%} & Human \\ \hline
\end{tabular}%
}
\caption{Human evaluation of fluency, relevancy, and answerability. We used the top-ranked 30\% of judges provided by a crowdsourcing service. Three judges performed each hit. Spammers were blocked at runtime. Agreement with most common was 81\% overall. MSQG refers to $\text{MSQG}_{sharedh,rmrep}$. The upper table shows evaluations of individual models and the lower shows pairwise comparisons: $(\textit{Human} \leftrightarrow \text{MSQG}_{sharedh,rmrep})$ and $(\text{MSQG}_{sharedh,rmrep} \leftrightarrow \text{S2S})$. Comparison results are significant at $p<0.00001$.}
\label{tab:humaneval}
\end{table*}

\begin{table}[]
\centering
\resizebox{\columnwidth}{!}
{%
\begin{tabular}{l|ccc|c}
\hline
 & \multicolumn{3}{c|}{\textbf{Retrieval Statistics}} &  \\
\multicolumn{1}{c|}{\textbf{Model}} & \textbf{MRR} & \textbf{MRR@10} & \textbf{nDCG}  & \multirow{-2}{*}{\textbf{\begin{tabular}[c]{@{}c@{}}Unique $\tilde{q}$\\ $\%$ dev.\end{tabular}}} \\ \hline
S2S & 0.0520  & 0.0266 & 0.2147 & 70.6 \\
S2S$_{rmrep}$ & 0.0540  & 0.0284 & 0.2152 & {\textbf{80.4}} \\
MESD & 0.0509  & 0.0248 & 0.2141 & 68.6 \\
MEMD$^{mult}$ & 0.0513  & 0.0256 & 0.2142 & 61.4 \\
MEMD & 0.0560  & 0.0287 & 0.2209 & 66.9 \\
MSQG$_{sharedh}$ & {\color[HTML]{000000}0.0569}  & {\color[HTML]{000000}0.0298} & {\color[HTML]{000000}0.2220} & 67.0 \\
MSQG$_{sharedh,rmrep}$ & {\textbf{0.0704}}  & {\textbf{0.0441}} & {\textbf{0.2337}} &  {\color[HTML]{000000}70.3} \\ \hline
\end{tabular}%
}

\caption{Our proposed model MSQG$_{sharedh,rmrep}$ significantly outperforms baselines, based on the automated retrieval statistics. Discussion of the proportion of unique questions is dealt in supplementary material.}
\label{tab:stats}
\end{table}

\subsection{Dataset}
We use the Microsoft MAchine Reading COmprehension Question-Answering Dataset (MS-MARCO-QA)  \citep{DBLP:conf/nips/NguyenRSGTMD16}, where a single data instance consists of an anonymized Bing search query $q$ and top-10 retrieved passages. Among the 10 passages, a passage is labelled \textit{is-selected:True} if annotators used it, if any, to construct answers, and most instances contain one or two selected passages. 
For training S2S, we use a single selected passage $p^* \in \{p_1, p_2, \dots, p_{10} \}$ as input, and the query $q$ as target output.

\subsection{Constructing Evaluation Sets}
For automatic evaluation, we follow the standard evaluation method from the MS-MARCO Re-Ranking task. For each generated question $\tilde{q}$, we construct an evaluation set that contains 100 passages in total.\footnote{A small number of evaluation sets had less than 100 passages because of duplicates between the source 10-passage set and the 90 passages retrieved via BM25.} 

First, using the 10-passage sets from the MS-MARCO-QA development dataset as inputs, we generate common questions with the baselines and our MSQG models, decoded for a maximum length of 25 words. A sample generation is provided in the SM.
Secondly, we evaluate the generations by using the pre-trained BERT-based MS-MARCO passage re-ranker $\mathcal{R}$, which is publicly available and state-of-the-art as of April 1, 2019 \citep{DBLP:journals/corr/abs-1901-04085}. 
We assess whether the 10-passage set used to generate the question ranks higher than 90 other passages drawn from a pool of $\sim$8.8 million MS-MARCO passages using the generated question. 
These 90 passages are retrieved via a different criterion: BM25 \citep{Robertson:2009:PRF:1704809.1704810} using Lucene\footnote{https://lucene.apache.org/}.
Note that there are multiple 10-passage sets that generate the same question $\tilde{q}$. For each of these 10-passage sets, we construct a 100-passage evaluation set using the same 90 passages retrieved via the BM25 criterion.


\subsection{Evaluation Metrics}\label{metrics}
\paragraph{MRR, MRR@10, nDCG }
An input to the re-ranker $\mathcal{R}$ is a concatenation of the generated question and one passage i.e. $[\tilde{q}, p]$. For each pair, it returns a score $\in (0,1)$ where 1 denotes that the input passage is the most suitable for $\tilde{q}$. We score all 100 pairs in an evaluation set. For the source 10-passage set, we average the 10 scores into one score as one combined document and obtain the retrieval statistics MRR, MRR@10 \cite{Voorhees:2001:TQA:973890.973895,radev-etal-2002-evaluating}, and nDCG \cite{Jarvelin:2002:CGE:582415.582418} (see the SM for details).

\paragraph{Human Judgments} 
We also conduct human evaluation where we compare questions generated by MSQG$_{sharedh,rmrep}$ and the S2S
baseline, and the reference question using three criteria: \textit{fluency}, \textit{relevancy}, and \textit{answerability} to the original 10 passages. We randomly select 200 (10-passage, reference question) sets from which we generate questions, yielding 2,000 (passage, question) evaluation pairs for our model, baseline, and reference, respectively (see the SM for details).

\subsection{Results}
Table~\ref{tab:stats} shows the mean retrieval statistics and their proportion of unique generated questions from 55,065 10-passage instances. Notice that our proposed MSQG models are more effective in terms of retrieving the source 10-passage sets. Particularly, MSQG$_{sharedh,rmrep}$ outperforms the baselines in all metrics, indicating that broadcasting the mean of the document vectors to initialize the decoders (\textit{sharedh}), and increasing the coverage of vocabulary (\textit{rmrep}) are effective mechanisms for generating common questions.

Overall, the retrieval statistics are relatively low. 
Most 100 passages in the evaluation sets have high pair-wise cosine similarities. We computed similarities of passage pairs for a significant portion of the dataset until convergence. 
A random set of 10 passages has an average pair-wise similarity of 0.80, whereas the top-10 re-ranked passages have an average of 0.85 based on BERT \citep{DBLP:journals/corr/abs-1810-04805} embeddings. 
Given the small similarity margin, the retrieval task is challenging. 
Despite of low statistics, we obtained statistical significance based on MRR with $p<0.00001$ between \textit{all} model pairs (see the SM for details).

Human evaluation results are shown in Table \ref{tab:humaneval}. In the comparison tasks, our proposed model significantly outperforms the strong baseline by a large margin. Nevertheless, judges preferred the reference over our model on all three aspects. The individual tasks corroborate our observations.

\section{Conclusion}\label{conclusion}
We present a new task of generating common questions based on shared concepts among documents, and extensively evaluated multi-source encoder-decoder framework models, including our variant model MSQG applied to this new task. We also provide an empirical evaluation framework based on automated metrics and human judgments to evaluated multi-source generation framework for generating common questions. 
\clearpage
\nocite{*}
\bibliographystyle{acl_natbib}
\bibliography{main}

\clearpage
\appendix

\section*{Supplementary Material}
\section{Full Experiment Results}

\begin{table}[]
\centering
\resizebox{\columnwidth}{!}
{%
\begin{tabular}{l|ccc|c}
\hline
 & \multicolumn{3}{c|}{\textbf{Retrieval Statistics}} &  \\ \cline{2-4}
\multicolumn{1}{c|}{\textbf{Model}} & \textbf{MRR}  & \textbf{MRR@10} & \textbf{nDCG}& \multirow{-2}{*}{\textbf{\begin{tabular}[c]{@{}c@{}}Unique $\tilde{q}$\\ $\%$ dev.\end{tabular}}} \\ \hline
S2S-$\mathcal{M}_{256}$ & {\color[HTML]{000000} 0.0368}  & {\color[HTML]{000000} 0.0147}& {\color[HTML]{000000} 0.1943} & 66.3 \\
S2S-$\mathcal{M}_{512}$ & 0.0393  & 0.0165 & 0.1980 & 71.1 \\
S2S & 0.0520  & 0.0266 & 0.2147 & 70.6 \\
S2S$_{rmrep}$ & 0.0540 & 0.0284 & 0.2152 & {\textbf{80.4}} \\
MESD-$\mathcal{M}_{512}$ & {\color[HTML]{000000} 0.0367}  & {\color[HTML]{000000} 0.0147} & {\color[HTML]{000000} 0.1941} & {\color[HTML]{000000} 72.6} \\
MESD & 0.0509  & 0.0248 & 0.2141 & 68.6 \\
MSQG-$\mathcal{M}_{512}$ & 0.0450  & 0.0210 & 0.2056 & 72.0 \\
MSQG$^{mult}$ & 0.0513 & 0.0256 & 0.2142 & 61.4 \\
MSQG & 0.0560 & 0.0287 & 0.2209 & 66.9 \\
MSQG$_{sharedh}$ & {\color[HTML]{000000} 0.0569}  & {\color[HTML]{000000} 0.0298} & {\color[HTML]{000000} 0.2220} & 67.0 \\
MSQG$_{sharedh,rmrep}$ & {\textbf{0.0704}}  & {\textbf{0.0441}} & {\textbf{0.2337}} & {\color[HTML]{000000} 70.3} \\ \hline
\end{tabular}%
}
\caption{Full results, comparing models constructed with $\mathcal{M}_{256}$, $\mathcal{M}_{512}$, and $\mathcal{M}^{attn}_{256}$. $\mathcal{M}_{512}$ has the most number of parameters among the three considered.}
\label{tab:stats}
\end{table}

Table~\ref{tab:stats} shows the retrieval results on a larger set of baselines and MSQG models. $\mathcal{M}^{attn}_{256}$ is an attention-based encoder-decoder with hidden size 256 for both encoder and decoder. $\mathcal{M}_{256}$ and $\mathcal{M}_{512}$ are non-attention encoder-decoders with hidden sizes 256 and 512. S2S denotes $\mathcal{M}^{attn}_{256}$, as in the main paper. It shows that models constructed using $\mathcal{M}^{attn}_{256}$ are more effective as opposed to models using $\mathcal{M}_{512}$ which has more parameters. Furthermore, we see that the averaging scheme in the \textit{Reduction} step, broadcasting the same encoder mean, and increasing coverage of vocabulary tokens are important features to generating common questions using MSQG models.

\section{Effect of Document Order on S2S}

To examine if the order of multiple input documents are critical for S2S, we obtain the attention weights at each decoding time, gathered across the development dataset. Next, we perform a simple ordinary least squares regression, where the predictors are indexed word positions in a concatenated input, and responses are assumed noisy attention weights over the development dataset for each word position. 

The slope coefficient fell within the 95\% confidence interval that includes the null: $\left[\num{-2.75e-5}, \num{3.03e-5}\right]$ and a statistically significant intercept value of $0.0021$. The result also validates that an average 10-passage string is approximately 476 ($\approx\frac{1}{0.0021}$) words long. Thus, we conclude that the attention weights are evenly distributed across multiple document at test time, and the document ordering is not critical to the performance of S2S.


\section{Clustering Duplicate 10-passage Sets}

In the MS-MARCO-QA dataset, there are many highly similar 10-passage sets retrieved from semantically close MS-MARCO queries. Examples of semantically close MS-MARCO queries include [\textit{``symptoms blood sugar is low", ``low blood sugar symptoms", ``symptoms of low blood sugar levels" ,``signs and symptoms of low blood sugar", ``what symptoms from low blood sugar", ... }], from which we expect duplicate generated questions, thus in sum, less than 55,065 different questions. 

\begin{figure}[]
\centering
\includegraphics[width=1.0\linewidth]{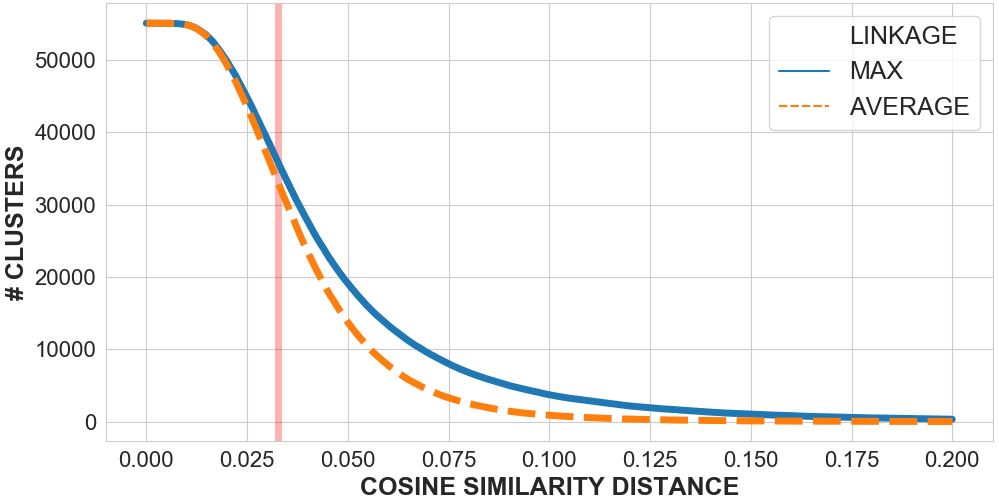} 
\caption{Agglomerative Clustering of 55,065 source 10-passage sets. Each set is represented by the mean of 10 BERT embeddings. Both max and average linkages yield the same inflection point at 0.0326, corresponding to 35,928 and 32,871 clusters. This method implies that the target proportion of unique generations should be at least 65\% or 60\%, which all models but MSQG$^{mult}$ achieve. }
\label{clustering}
\end{figure}

Therefore, to estimate the \textit{target} proportion of unique generations, we examine the number of semantically similar 10-passage sets through agglomerative clustering. Figure~\ref{clustering} shows cluster results with varying degrees of affinity thresholds, and observe that the effective models should generate at least 65\% unique questions from the development dataset. This, together with the low retrieval statistics of MSQG$^{mult}$, implies that multiplying the distributions is not an appropriate \textit{Reduction} step. 


On the other hand, generating the most number of unique questions does not imply that the model better generates common questions. In particular, S2S$_{rmrep}$ generates the most diverse questions, however, its retrieval statistics are significantly lower than its MSQG counterparts.

\section{Statistical Significance Tests}

Retrieval evaluation on $\sim$55K evaluation sets using the re-ranker $\mathcal{R}$ is compute-intensive. Thus, for each model, we randomly sample and obtain retrieval statistics for 15K evaluation sets which are enough to mimic the true evaluation set distribution. 

Then, to assess statistical significance, we use a non-parametric two-sample test, such as Mann-Whitney (MW) or Kolmogorov-Smirnov statistic, and test whether any pair of 15K retrieval sets between two models come from the same distribution. In our task, both tests reached the same conclusion. MW two-sample tests on MRR results showed statistical significance at $p<0.00001$ for \textit{all} model pairs dealt in the main paper, in spite of the relatively low retrieval statistics.



\section{Human Evaluation Templates}

UHRS comparison and individual task instructions and shown in the next pages.






\section{Generated Questions Sample}
\setlength\parindent{0pt}
\textbf{Passage 1:} \textit{cucumbers and zucchini look similar but have nutritional differences . photo credit martin poole / digital vision / getty images . do n't let the similarities between cucumbers and zucchini confuse you . even though both cylindrical vegetables are dark green with white flesh , they are distinctively different species . both cucumbers and zucchini belong to the curcurbit family , which also counts gourds , melons , pumpkins and squash among its members . cucumbers and zucchini differ both in how people commonly eat them and in their nutritional values . people almost always eat cukes raw , while zucchini is more often cooked .}
\\\\
\textbf{Passage 2:} \textit{cucumber and squash seedlings both have elongated foliage for the first set of leaves after they emerge from the soil . the second set of leaves on a seedling varies . cucumber leaves are in the shape of a triangle and are flat in the center and rough to the touch . squash plants vary in shape as to the particular variety , but have three to five lobes and are larger than cucumber leaves . zucchini squash has elongated serrated leaves .}
\\\\
\textbf{Passage 3:} \textit{zucchini vs cucumber . zucchini and cucumber are two vegetables that look mightily similar and hard to distinguish from each other . but in close inspection , they are actually very different . so read on . zucchini . zucchini is defined to be the kind of vegetable that is long , green colored and has many seeds .}
\\\\
\textbf{Passage 4:} \textit{as a general rule , we prefer cucumbers raw and zucchini cooked . while you ca n't replace one with the other , zucchinis and cucumbers do complement one another . slice two cucumbers , two zucchinis and one sweet onion , and soak them all in rice vinegar for at least an hour in the refrigerator .}
\\\\
\textbf{Passage 5:} \textit{cucumber and zucchini are popular vegetables that are similar in appearance and botanical classification . but they differ significantly in taste , texture and culinary application . zucchini and cucumber are both members of the botanical family cucurbitaceae , which includes melons , squashes and gourds .}
\\\\
\textbf{Passage 6:} \textit{melon vs. squash . the cucumber is not particularly sweet , but it shares a genus with the cantaloupe and is botanically classified as a melon . the zucchini is a variety of summer squash and is of the same species as crookneck squash .}
\\\\
\textbf{Passage 7:} \textit{cucumber vs. zucchini . side by side , they might fool you : cucumbers and zucchinis share the same dark green skin , pale seedy flesh , and long cylindrical shape . to the touch , however , these near - twins are not the same : cucumbers are cold and waxy , while zucchinis are rough and dry . the two vegetables also perform very differently when cooked .}
\\\\
\textbf{Passage 8:} \textit{the second set of squash leaves grow much quicker and larger than cucumber leaves in the same time . squash leaves may be up to four times as large as a cucumber leaf when they are the same age .}
\\\\
\textbf{Passage 9:} \textit{in reality , zucchini is really defined as a vegetable so when it comes to the preparation of it , it has different temperament . cucumber . cucumber is both classified as a fruit and a vegetable . it is long and is green in color , too . it is part of what they call the gourd family .}
\\\\
\textbf{Passage 10:} \textit{zucchini 's flowers are edible ; cucumber 's flowers are not . zucchini is generally considered as a vegetable ; cucumber is classified as both a fruit and a vegetable . yes , they can fool the eye because of their similar look but as you go deeper , they are very different in so many ways .}
\\\\
\textbf{Question generated by MSQG$_{sharedh,rmrep}$:} 

\textit{what are the difference between cucumber and zucchini}
\\\\
\textbf{Question generated by S2S:} 

\textit{different types of zucchini}
\\\\
\textbf{Reference question:} 

\textit{difference between cucumber and zucchini}

\begin{figure*}[]
\centering
\includegraphics[width=1.0\linewidth]{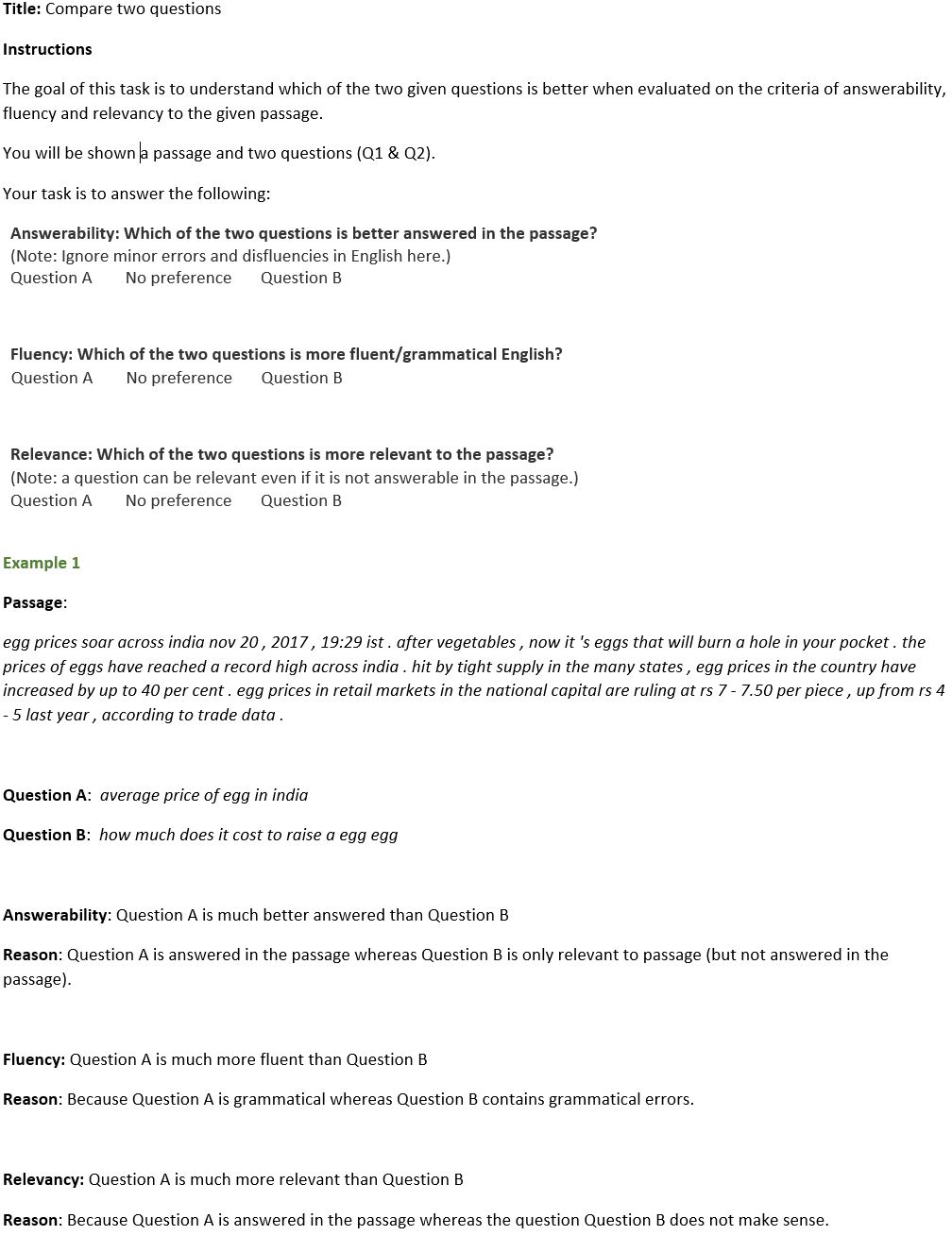} 
\label{comparison}
\end{figure*}

\begin{figure*}[]
\centering
\includegraphics[width=1.0\linewidth]{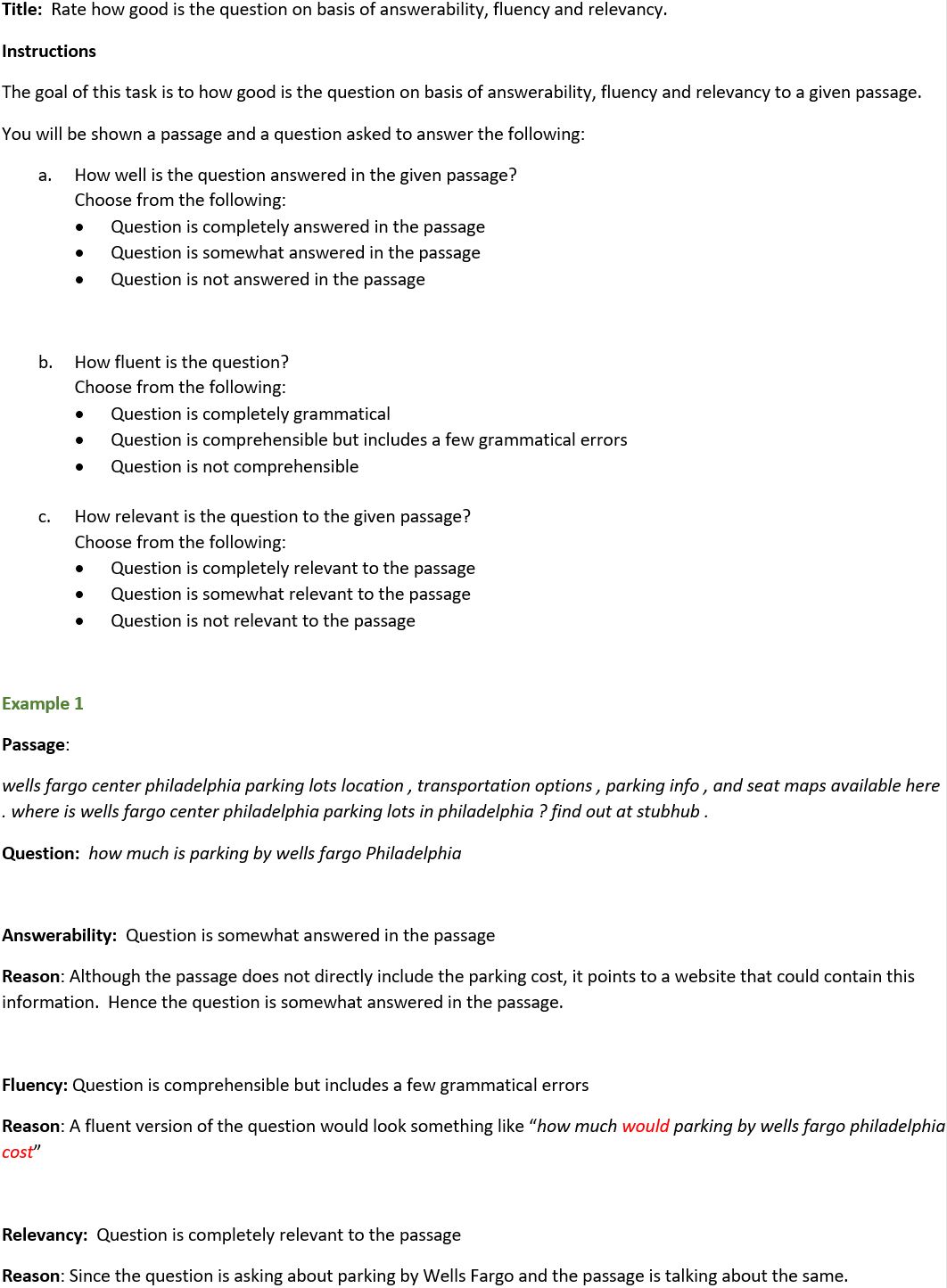} 
\label{comparison}
\end{figure*}

\end{document}